\documentclass[conference,compsoc]{IEEEtran}
\IEEEoverridecommandlockouts

\usepackage{cite}
\usepackage{amsmath,amssymb,amsfonts}
\usepackage{algorithmic}
\usepackage{graphicx}
\usepackage{textcomp}
\usepackage{xcolor}
\usepackage{subcaption}
\usepackage{balance}

\def\BibTeX{{\rm B\kern-.05em{\sc i\kern-.025em b}\kern-.08em
    T\kern-.1667em\lower.7ex\hbox{E}\kern-.125emX}}
\begin{document}

\title{Tracing the Interactions of Modular CMA-ES Configurations Across Problem Landscapes}

\author{\IEEEauthorblockN{Ana Nikolikj}
\IEEEauthorblockA{\textit{Computer Systems Department} \\
\textit{Jožef Stefan Institute}\\
Ljubljana, Slovenia\\
ana.nikolikj@ijs.si}
\and
\IEEEauthorblockN{Mario Andrés Muñoz}
\IEEEauthorblockA{\textit{Computing and Information Systems}\\ \textit{The University of Melbourne} \\
Parkville, VIC 3010 Australia \\
munoz.m@unimelb.edu.au}
\and
\IEEEauthorblockN{Eva Tuba}
\IEEEauthorblockA{\textit{Trinity University} \\
San Anotnio, TX \\
\textit{Singidunum University}\\
Belgrade, Serbia\\
etuba@ieee.org}
\and
\IEEEauthorblockN{Tome Eftimov}
\IEEEauthorblockA{\textit{Computer Systems Department} \\
\textit{Jožef Stefan Institute}\\
Ljubljana, Slovenia\\
tome.eftimov@ijs.si}
}

\maketitle

\IEEEpubidadjcol

\begin{abstract}
This paper leverages the recently introduced concept of algorithm footprints to investigate the interplay between algorithm configurations and problem characteristics. Performance footprints are calculated for six modular variants of the CMA-ES algorithm (modCMA), evaluated on 24 benchmark problems from the BBOB suite, across two-dimensional settings: 5-dimensional and 30-dimensional. These footprints provide insights into why different configurations of the same algorithm exhibit varying performance and identify the problem features influencing these outcomes. Our analysis uncovers shared behavioral patterns across configurations due to common interactions with problem properties, as well as distinct behaviors on the same problem driven by differing problem features. The results demonstrate the effectiveness of algorithm footprints in enhancing interpretability and guiding configuration choices.
\end{abstract}


\section{Introduction}
\label{sec:intro}

Black-box optimization (BBO) involves developing and analyzing algorithms to tackle problems where the objective function is not explicitly accessible, relying only on iterative sampling and evaluation of candidate solutions~\cite{bajaj2021black}. In single-objective optimization (SOO), the aim is to identify the optimal solution from a set of candidates for a specific problem instance. A problem instance is defined by its variables outlining the solution space, an objective function for assessing solution quality, and a similarity measure between solutions, known as the neighborhood. Continuous optimization focuses on cases where these variables are real-valued, and the neighborhood is considered Euclidean. The process entails using an optimization algorithm to generate candidate solutions, which are evaluated by the objective function until convergence. These algorithms often introduce randomness by sampling from probability distributions over the problem space. Meta-heuristics, widely used in such contexts, offer resilient solutions even with limited computational resources or incomplete information~\cite{molina2018insight}. They are typically divided into population-based and single-solution-based approaches.

Various algorithms have been proposed for continuous SOO, and their effectiveness is often assessed through statistical analyses, reporting average performance across a set of benchmark problems~\cite{molina2018insight}. The algorithmic landscape encompasses various families, including covariance matrix adaptation evolution strategy (CMA-ES)~\cite{varelas2018comparative}, differential evolution (DE)~\cite{pant2020differential}, and particle swarm optimization (PSO)~\cite{kennedy1995particle}. For example, each class of evolutionary algorithms incorporates specific mechanisms like mutation, crossover, selection strategies, and configurable hyperparameters. Algorithm instances representing different configurations of a given algorithm class may perform differently depending on the problem. Automated configuration techniques~\cite{lopez2016irace,bartz2010sequential} and hyper-heuristics~\cite{drake2020recent} are employed to identify the best-performing configurations (the optimal combination of components or hyperparameters or both). These methods explore the relationship between algorithm performance and a predefined set of training problems, enabling dynamic selection and fine-tuning of algorithms.

In this direction, rather than analyzing algorithm configurations in isolation, the approach proposed in~\cite{de2021tuning} advocates using a standardized modular optimization framework. This framework enables systematic evaluation of different algorithm configurations while ensuring consistent implementation across all variants. The core concept involves breaking down an algorithm into smaller, independent units known as modules. Each module offers configurable options that influence the algorithm’s behavior and can be combined to form new algorithm variants. The studies highlight the benefits of modular frameworks~\cite{van2018towards,vermetten2023modular,camacho2021pso}, including more equitable implementation-based comparisons and the ability to explore interactions between different modules.

Although techniques from machine learning (ML), such as functional ANOVA (fANOVA and its extensions)~\cite{hutter2014efficient} and Shapley values~\cite{sundararajan2020many} from game theory, have been utilized to explain the influence of hyperparameters or modules by examining their impact on performance, notable challenges remain~\cite{nikolikj2024quantifying,van2024explainable}. Moreover, ablation studies were conducted to assess the impact of individual modules on performance across different problems, without exploring their properties in depth~\cite{van2017algorithm}. These explainable approaches have not fully bridged the gap in understanding the interaction between hyperparameters or modules and the properties of optimization problems. A recent approach was introduced to visually explore the relationships between algorithm performance, parameter settings, and characteristics of the problem landscape~\cite{rasulo2024extending}. This entails jointly analyzing a 2D instance space, where problem instances are projected, and a 2D configuration space, where algorithm configurations are mapped. Building on the dimensionality reduction approach used in Instance Space Analysis, an optimization problem has been formulated to derive projections for these spaces, ensuring an interpretable relationship between them. This highlights the ongoing need for deeper insights into the interplay between algorithm configurations and problem properties.

\noindent\textbf{Our contribution}: We utilize a recently introduced approach known as algorithm footprints~\cite{nikolikj2023algorithm,nikolikj2024comparing} to explore the relationships between algorithm configurations and problem characteristics. Specifically, we generate algorithm configuration footprints for six modular CMA-ES (modCMA) configurations~\cite{van2018towards},  tested on 24 problems from the BBOB benchmark suite~\cite{hansen2021coco}, with separate evaluations in 5 and 30 dimensions. These footprints help explain why different configurations of the same algorithm yield varying performance outcomes and point out the problem characteristics driving those outcomes for each problem. Furthermore, the analysis highlights similar behavior across different configurations caused by shared interactions with problem properties and distinct behavior on the same problem due to the influence of different problem characteristics. 

Section~\ref{sec:rw} provides an overview of relevant studies. 
Section~\ref{sec:footprint} briefly explains the benchmarking footprints methodology. Section~\ref{sec:exp_design} outlines the experimental setup, while Section~\ref{sec:results} presents the key results. Section~\ref{sec:discussion} provides an interpretation of the findings and highlights future directions. Lastly, Section~\ref{sec:conclusion} concludes the paper.

\section{Related work}
\label{sec:rw}
Early efforts in this area focused on a detailed evaluation of module significance in CMA-ES by analyzing the performance contributions of individual modules through an enumeration of all possible module options~\cite{van2017algorithm}. However, this approach becomes impractical as the number of modules grows.

In~\cite{de2021tuning} and~\cite{vermetten2023modular} made use of the irace algorithm‐tuning framework~\cite{lopez2016irace} to systematically survey a vast configuration space and uncover elite configurations—that is, the top‐performing algorithm variants across a suite of optimization problems. Module importance was quantified based on the frequency of each module's inclusion in these elite configurations. This iterative process involved progressively expanding the module space explored by irace~\cite{lopez2016irace}. As additional modules were added, their interplay became more complex—an effect made visible through frequency‐based plots of the elite configurations. Yet, a detailed quantitative breakdown that teases apart each module’s contributions -whether individual, pairwise, or higher‐order - has not yet been performed.

Recent research has investigated how different modules interact and influence configuration performance for modular CMA and modular Differential Evolution (modDE) using techniques like functional ANOVA and Shapley values~\cite{nikolikj2024quantifying,van2024explainable}. These approaches advance the analysis by measuring the impact of modules individually and their combinations (e.g. pairs and triplets) on overall performance. However, these analyses are limited to the algorithm performance space and fail to account for the characteristics of the given problems.
\section{Algorithm configuration footprints}
\label{sec:footprint}
Workflow of the recently proposed benchmarking algorithm footprints methodology~\cite{nikolikj2024comparing, NIKOLIKJ2025101895} includes:

\noindent\textbf{(1) Training a meta-model}: A multi-target regression (MTR) model~\cite{korneva2020towards} is trained to predict algorithm performance based on the landscape features of the training problem instances. The MTR model leverages the capacity to predict multiple algorithm outcomes from shared features.

\noindent\textbf{(2) Meta-Representation Generation}: Apply explainability methods (e.g., SHAP) to calculate local feature importance for test instances, generating meta-representations that connect landscape features to algorithm performance. Each problem instance, paired with a specific algorithm, will have a unique meta-representation reflecting the relationship between its landscape features and its performance.

\noindent\textbf{(3) Meta-Representation Clustering}: Cluster meta-representations to identify regions with varying algorithm performance caused by different problem landscape features interactions. Rank clusters by their average performance, from lowest to highest.

\noindent\textbf{(4) Footprint Comparison}: Compare the distribution of meta-representations across clusters for different algorithms on the same instance to uncover similarities, differences, and challenging regions for the portfolio.

\noindent\textbf{(5) Feature Analysis}: Identify critical landscape features influencing problem difficulty.

In this study, we applied the methodology to a portfolio of algorithms from the same class but with different configurations. By tracing their interactions with problem landscapes, this approach helps explain how variations in configurations (hyperparameters or modules) of a core algorithm lead to differing performance outcomes.

\section{Experimental design}
\label{sec:exp_design}

\noindent\textbf{Performance data}: We leveraged on performance results from the earlier study by~\cite{kostovska2024using}, which evaluated 324 modular CMA-ES configurations on the BBOB benchmark suite, running five instances per problem at dimensions $d=\left\{5,30\right\}$. We selected six configurations per dimension (Table~\ref{tab:hyperparameters_modcma_5d} for 5$d$), including the best and worst performers (based on average performance) along with four standard CMA-ES variants, including elitist and local restart configurations. Performance was measured using the fixed-budget metric, assessing solution quality after 
1500$d$ function evaluations.

\noindent\textbf{Problem suite}: We employed 24 noiseless, single-objective black-box optimization problems from the BBOB benchmark suite ~\cite{hansen2021coco}. Each problem has multiple instances generated through scaling, shifting, and rotation transformations. For this study, we used the first five instances of each problem for $d=\left\{5,30\right\}$, resulting in two suites of 120 instances each.

\noindent\textbf{Landscape features}: Exploratory Landscape Analysis (ELA) feature data is reused from previous studies~\cite{quentin_renau_2020_3886816,kostovska2024using}. The total of 46 features were computed using Sobol sampling with a sample size of $100d$ across 100 independent repetitions, and the median feature values were used to represent each problem instance. We applied forward feature selection to identify the most relevant features among the 46 available.

\noindent\textbf{MTR models}: We evaluated three machine‐learning approaches to select a high‐performing MTR model:Multi-Task Elastic Net (MTEN)~\cite{zou2005regularization}, Random Forest (RF)~\cite{biau2016random}, and Neural Network (NN)~\cite{glorot2010understanding}. RF and MTEN were implemented using \textit{scikit-learn}~\cite{pedregosa2011scikit}, while NN utilized the \textit{keras} package~\cite{chollet2015keras}.Before training, we applied min–max normalization to the ELA features, mapping them into the $[0,1]$ interval. The normalization parameters were determined from the training set and then reused to transform the test set, ensuring consistency.

\noindent\textbf{MTR hyper-parameter tuning}: We performed hyperparameter optimization using Optuna’s Python implementation~\cite{akiba2019optuna} of the Tree-structured Parzen Estimator (TPE) sampler to pinpoint the best configuration. The process involves a budget of 50 trials: 40 trials utilize Random Search to explore the search space broadly, followed by 10 trials where TPE focuses on the most promising hyperparameter regions. The configuration whose hyperparameters maximize the cross-validation score on the training data is adopted as the optimal setup.

\noindent\textbf{Model evaluation}: The dataset comprises 120 instances - five instances of each of the 24 BBOB problem classes. One instance from each class (24 in total) is set aside as the test set, leaving 96 instances for training. We then carry out 4-fold cross-validation on the training data, stratifying folds by problem class. Each model is trained and validated across the four folds, and the average performance determines the best model. That model is finally retrained on all 96 training instances and evaluated against the 24 held-out test instances.

Model performance is assessed using Mean Absolute Error (MAE) and R-squared (R$^2$). MAE measures the average magnitude of the errors between predictions and actual values, while R² quantifies the fraction of variance in the true outcomes that the model explains, indicating their degree of alignment.

\noindent\textbf{SHAP explanations as meta-features:} Using the SHAP library~\cite{NIPS2017_7062}, we compute problem instance‐level feature contributions from the selected ML model, yielding an \(n\)-dimensional SHAP meta‐feature vector \[ \mathbf{s} = \bigl(s_1, s_2, \dots, s_n\bigr), \] where each \(s_i\) denotes the importance of feature \(i\) to the algorithm performance.

\noindent\textbf{Clustering:} Hierarchical clustering~\cite{Mllner2011ModernHA} was applied to the meta‐representations of problem instances to form groups automatically. Cluster quality was evaluated via the Silhouette coefficient—higher values correspond to more well‐defined clusters—and we tuned hyperparameters by comparing Euclidean and cosine distance metrics, ultimately choosing the configuration that maximized the Silhouette score.

 \begin{table}[!t]
 \centering
 \caption{The hyperparameter settings for the six modCMA configurations in 5$d$.}
 \resizebox{.5\textwidth}{!}{
 \begin{tabular}{lrlllll}
 \hline
 Name & elitist & mirrored & base\_sampler & weights\_option & local\_restart \\
 \hline
Default & False & Off & gaussian & default & Off\\
 Elitism &  True & Off & gaussian & default & Off \\
 Mirrored sampling &  False & mirrored & gaussian & default & Off \\
 Local restart &  False & Off & gaussian & default & IPOP \\
 Best on average &  False & mirrored & Sobol & default & BIPOP \\
 Worst on average &  False & pairwise & Halton & equal & Off \\
 \hline
 \end{tabular}
}
 \label{tab:hyperparameters_modcma_5d}
 \vspace{-5mm}
 \end{table}

\begin{figure}[!hbt]
    \centering
    \begin{subfigure}[b]{0.37\textwidth}
        \includegraphics[width=\linewidth]{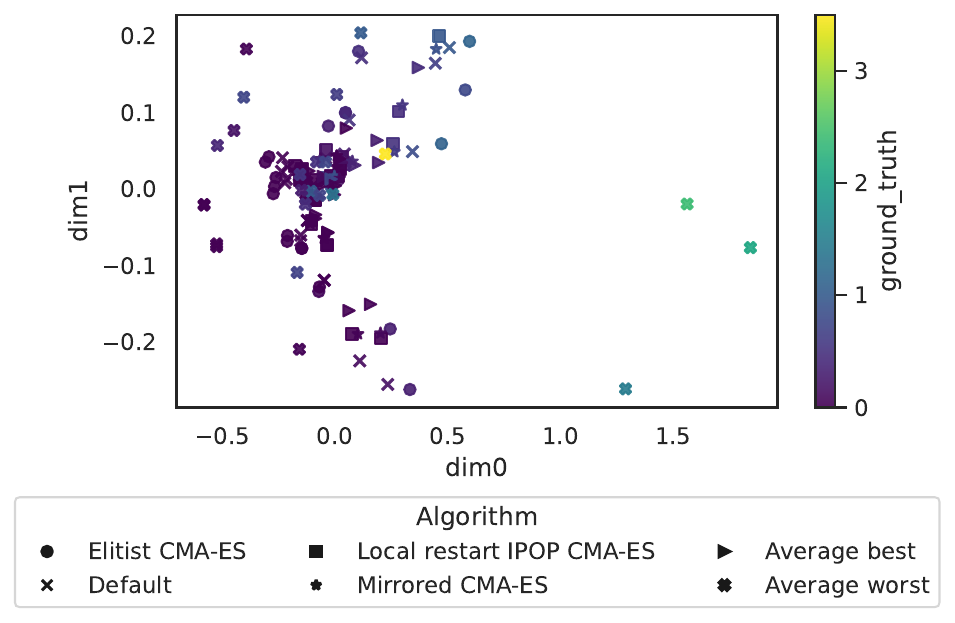}
        \caption{}
        \label{fig:ground_truth_2D_5d_modcma}
    \end{subfigure}
   
    \begin{subfigure}[b]{0.50\textwidth}
        \includegraphics[width=\linewidth]{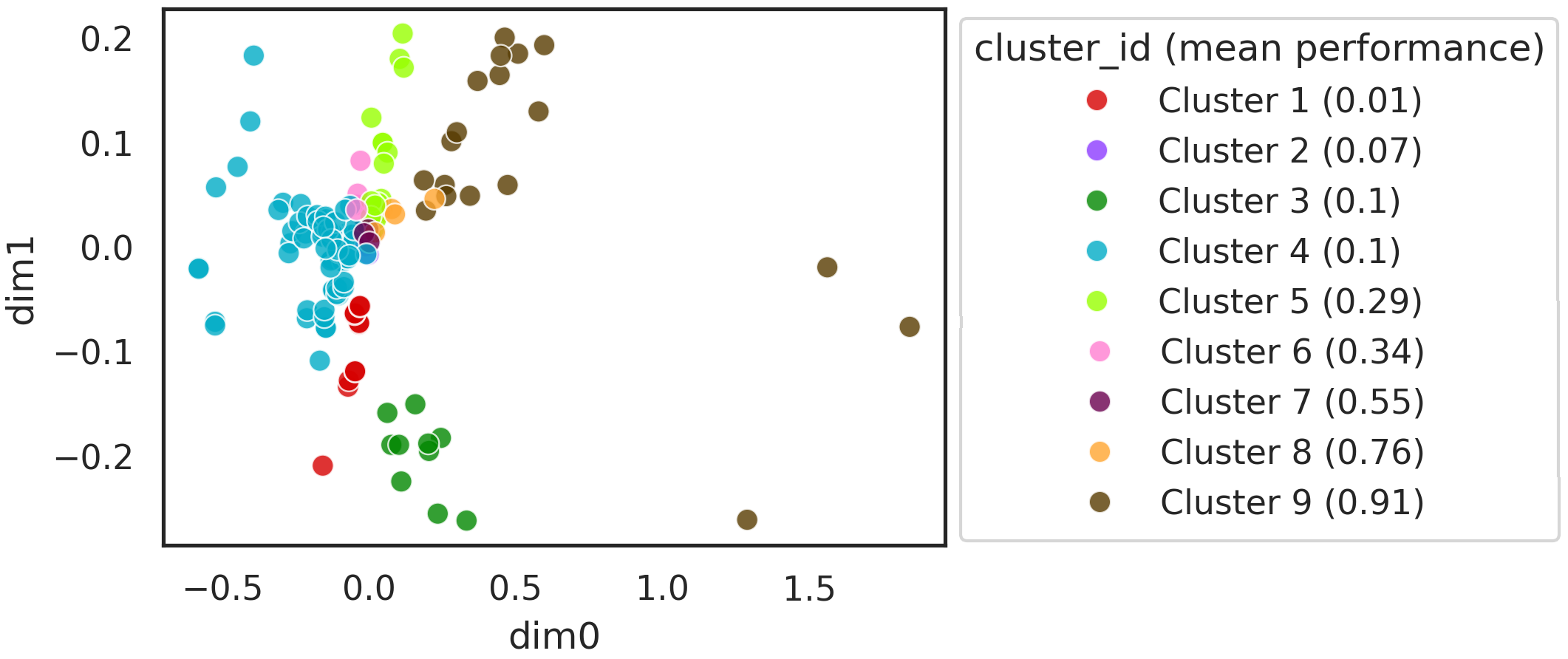}
        \caption{}
        \label{fig:clusters_2D_5d_modcma}
    \end{subfigure}
    \caption{The problem instances represented by the SHAP meta-features are projected into a two-dimensional vector space, with point colors showing: (a) the true performance across the six modCMA-ES configurations; and (b) the cluster labels assigned to the five-dimensional (5$d$) test instances.}
    \label{fig:results_modcma_5d}
     \vspace{-5mm}
\end{figure}

\section{Results}
\label{sec:results}

\begin{figure*}[!bht]
    \centering
    \includegraphics[trim=0.0cm 0.0cm 0.0cm 0.0cm, clip=true, scale=0.15]{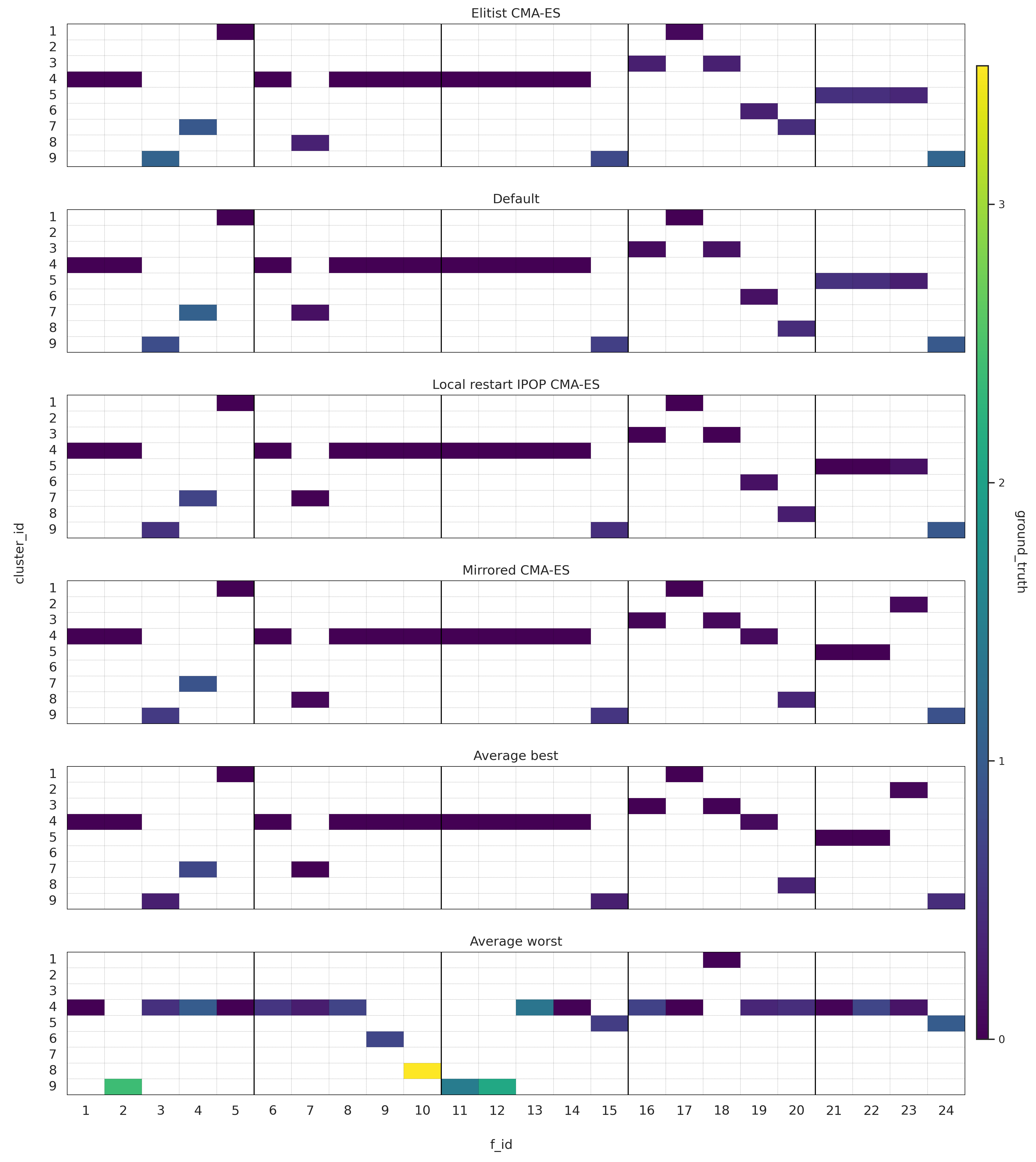}
    \label{fig:contagency_matrix_modcma_5d}
    \caption{For each modCMA configuration, the 5$d$ problem‐instance footprint is displayed, with the y‐axis marking the footprint regions and the x‐axis representing the problem class ID  ($f\_id$))}
    \label{fig:results_modcma_5d_footprint}
\end{figure*}

\begin{figure*}[!thb]
    \centering
    \begin{subfigure}[b]{0.25\textwidth}
        \includegraphics[width=\textwidth]{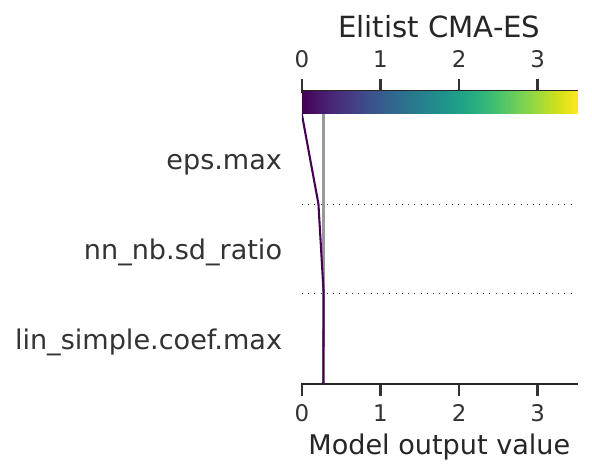}
        \caption{}\label{fig:modcma_0_5d_2}
    \end{subfigure}
    \begin{subfigure}[b]{0.25\textwidth}
        \includegraphics[width=\textwidth]{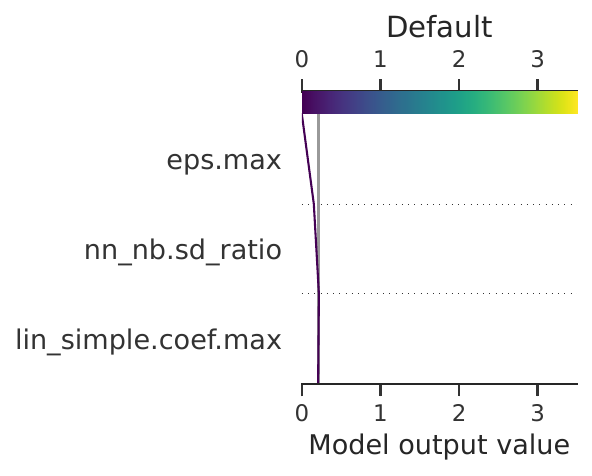}
        \caption{}\label{fig:modcma_162_5d_2}
    \end{subfigure}
    \begin{subfigure}[b]{0.25\textwidth}
        \includegraphics[width=\textwidth]{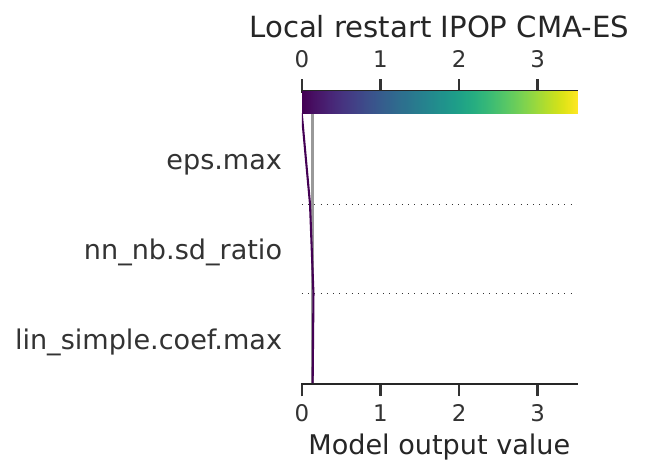}
        \caption{}\label{fig:modcma_164_5d_2}
    \end{subfigure}
    \begin{subfigure}[b]{0.25\textwidth}
        \includegraphics[width=\textwidth]{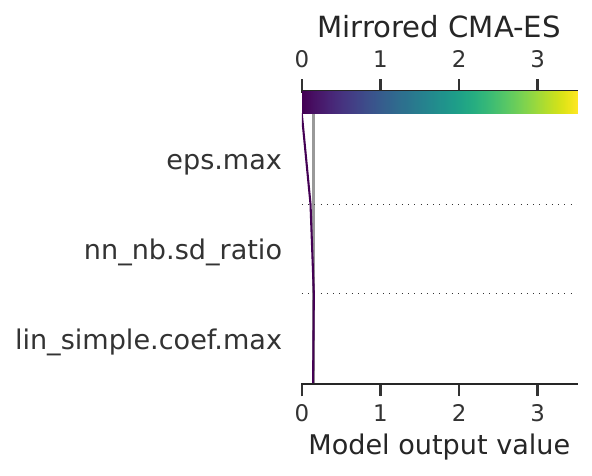}
        \caption{}\label{fig:modcma_216_5d_2}
    \end{subfigure}
    \begin{subfigure}[b]{0.25\textwidth}
        \includegraphics[width=\textwidth]{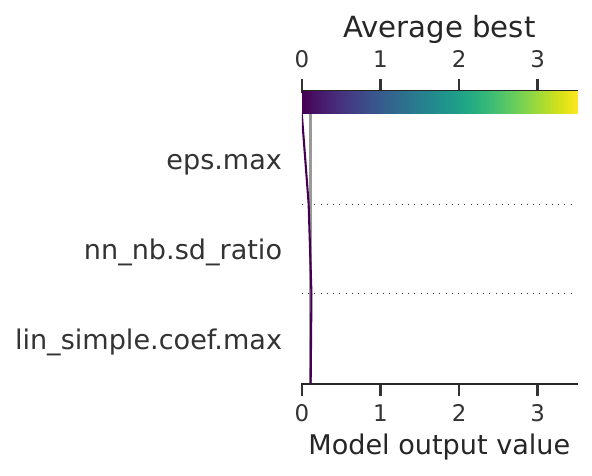}
        \caption{}\label{fig:modcma_238_5d_2}
    \end{subfigure}
    \begin{subfigure}[b]{0.25\textwidth}
        \includegraphics[width=\textwidth]{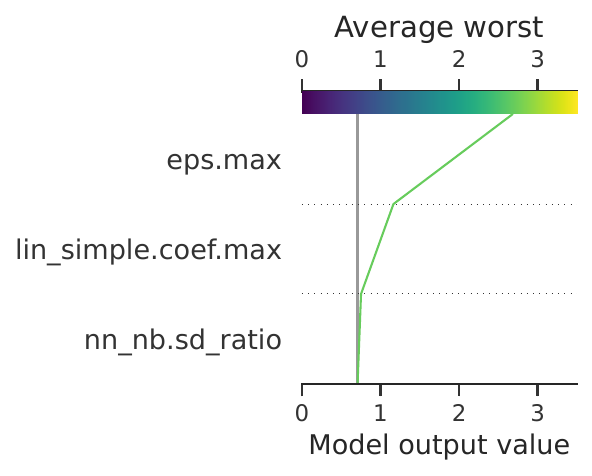}
        \caption{}\label{fig:modcma_312_5d_2}
    \end{subfigure}
    \caption{The subfigures depict the feature‐importance profiles for second problem class across different modCMA-ES configurations on five-dimensional instances.}
    \label{fig:post_hoc_patterns_modCMA_5d}
\end{figure*}

Figures \ref{fig:results_modcma_5d}–\ref{fig:post_hoc_patterns_modCMA_5d} present the footprint analyses for six modCMA configurations on 5$d$ problems. In particular, Figure~\ref{fig:results_modcma_5d} illustrates (a) the ground-truth performance and (b) the resulting clusters of the problem SHAP meta-features. Most problem instances are well-solved by all modCMA configurations (Figure~\ref{fig:ground_truth_2D_5d_modcma}), except for a few challenging ones for the poorest performer (as shown in the color map, where lower values denote better performance.). The clustering reveals nine distinct performance regions, driven by varying interactions among landscape features (Figure~\ref{fig:clusters_2D_5d_modcma}). 
We can also conclude that most of the performance across the three algorithms falls within the fourth cluster, indicating that similar interactions of landscape features are driving this outcome. However, we can also observe that similar performance may appear in nearby clusters, driven by different landscape feature interactions. For example, instances in the first two clusters show similar performance but differ due to distinct landscape-feature interactions affecting algorithm predictions. This also holds for other clusters, such as the 3rd and 4th, where identical performance stems from different landscape dynamics.


\begin{figure*}[htbp]
    \centering
    \begin{subfigure}[t]{0.25\textwidth}
        \centering
        \includegraphics[width=\linewidth]{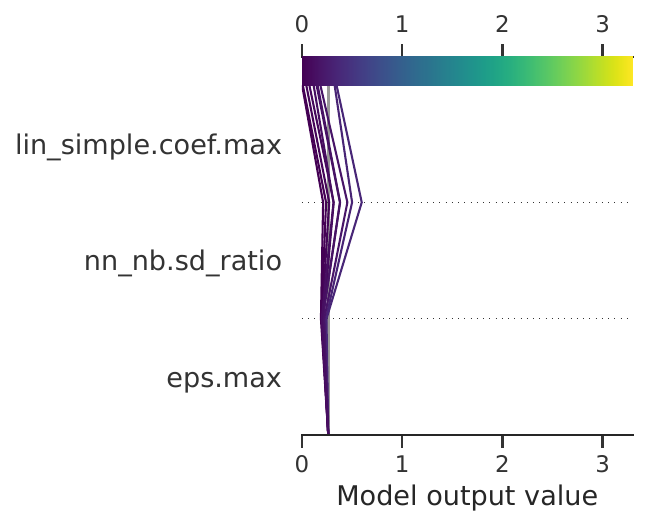}
        \caption{Third cluster.}
        \label{fig:sub1}
    \end{subfigure}
    \hfill
    \begin{subfigure}[t]{0.25\textwidth}
        \centering
        \includegraphics[width=\linewidth]{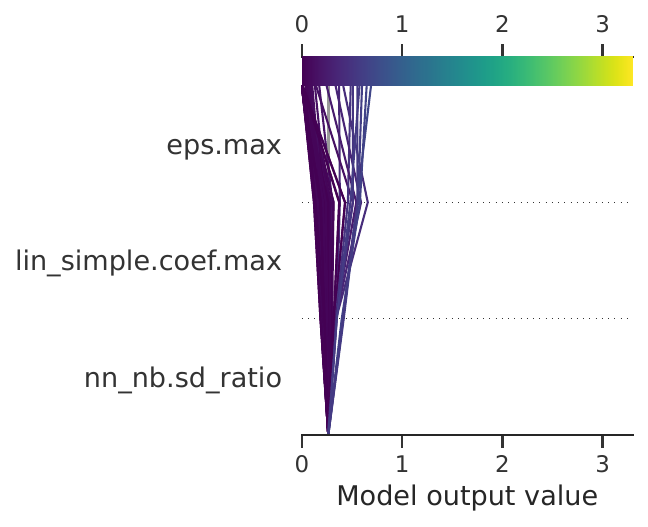}
        \caption{Fourth cluster.}
        \label{fig:sub2}
    \end{subfigure}
    \hfill
    \begin{subfigure}[t]{0.25\textwidth}
        \centering
        \includegraphics[width=\linewidth]{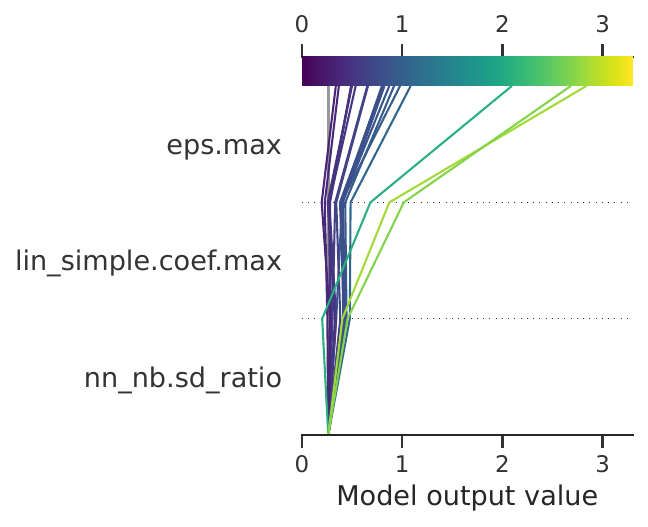}
        \caption{Ninth cluster.}
        \label{fig:sub3}
    \end{subfigure}
    \caption{Feature importance patterns in the third, fourth, and ninth performance regions.}
    \label{fig:perform_regions}
\end{figure*}

The clustering results, shown in Figure~\ref{fig:results_modcma_5d_footprint}, highlight the footprints of the modCMA configurations. Figure~\ref{fig:results_modcma_5d_footprint} shows a ``coverage matrix'' with columns representing problem instances and rows indicating the algorithm and its corresponding cluster. The colors reflect the ground-truth performance on each algorithm instance, with clusters ordered by ascending average performance. Comparing the clusters assigned to the same problem instances across different algorithms reveals similarities or differences in their behavior. Assigning a problem instance to the first cluster under one algorithm and to the last cluster under another highlights a pronounced performance discrepancy between them. The first algorithm tackles the instance successfully, while the second finds it challenging. Furthermore, these similarities and differences can be quantified using similarity measures, such as cosine similarity applied to the problem instance meta-features. Consistent with expectations, all configurations exhibit nearly identical footprint patterns, with the exception of the poorest-performing variant, which corresponds to the minimal differences observed in their ground-truth performance metrics in the raw data.

For a detailed examination, we centered our analysis on the problem class with ($f\_id=2$). Figure~\ref{fig:post_hoc_patterns_modCMA_5d} presents the dominant landscape features and their interrelations for this problem across every modCMA-ES variant, illustrating the interaction patterns that enable successful optimization in the five top-performing configurations and those that hinder the poorest-performing one. Each sub-figure displays a SHAP \textit{decision plot}, showing each feature's contribution (SHAP value) to the model's prediction. The features on the y-axis are ordered by descending importance, with the topmost feature exerting the greatest influence. The x-axis displays the model’s predicted value, and each curve shows how feature contributions accumulate from the baseline (mean ground-truth performance) to the final prediction. Curve colors encode predicted performance, ranging from dark blue (best) to yellow (worst). From this, it is evident that the five high-performing modCMA configurations share very similar contribution patterns, whereas these same feature interactions render the second problem especially difficult for the lowest-performing configuration, placing it in the poorest-performing region. This problem corresponds to the separable ellipsoidal function, and the set of modules used in the worst-performing configuration (see Table~\ref{tab:hyperparameters_modcma_5d}) fail to solve it. This problem fits the description of being unimodal and highly ill-conditioned. This would mean that we would expect huge values of lin\_simple.coef.max (in the order of $10^6$), and ic.eps.max, while values close to one for nn\_nb.sd\_ratio. 
A recent fANOVA study~\cite{nikolikj2024quantifying} revealed that the ``weights option'' module plays a significant role in performance for that problem, both in isolation and through interactions with other modules. However, this is true only when the option is set to ``default" or ``lambda"; in our case, it is configured as ``equal" weights. Additionally, another crucial module for this problem is the ``mirrored" module~\cite{nikolikj2024quantifying}. The ``Off" and ``mirrored" options yield better results, while the ``pairwise" option, used in our case, reduces performance. The combination of ``pairwise" and ``equal" further contributes to the worst performance. 
Additionally, as shown in Figure~\ref{fig:post_hoc_patterns_modCMA_5d}, the five best-performing configurations derive their most significant contribution from the eps.max feature, which enhances prediction accuracy, while the other two features contribute only marginally, staying close to the baseline. For the worst-performing configuration, both eps.max and lin\_simple.coef.max contribute to predicting its poor performance, making the problem particularly challenging for the selected module options.

The worst-performing configuration also struggled with the ill-conditioned uni-modal problems (10, 11, and 12). The fANOVA study~\cite{nikolikj2024quantifying} confirms these patterns, as these problems are grouped with the second problem based on the contribution of the modCMA modules. Similar module option patterns, as observed in the second problem, are also evident here. 
While the previous study focused solely on the performance space, our approach also links these module interactions to landscape properties. For instance, eps.max identifies ill-conditioning by estimating the highest slope found, while lin\_simple.coef.max reflects the highest coefficient in a linear model, providing similar insights. These features are shared by f10 (the rotated version of f2), f11 (the discuss function), f12 (the bent cigar function). They are unimodal with high condition numbers but with slight variations, e.g., the bent cigar is almost an ellipsoidal if the optimum is translated to the edges.

Due to space constraints, we cannot provide a detailed analysis of all problems. Instead, Figure~\ref{fig:perform_regions} illustrates feature importance patterns in the third, fourth, and ninth performance regions. Each subplot contains problem instances paired with a configuration that belongs to that cluster (see Figure~\ref{fig:clusters_2D_5d_modcma}). It is evident that the performance of configurations on problem instances in the third and fourth clusters is similar; however, these results are driven by different orders and interactions of key features. The ninth cluster represents the performance of configurations on problem instances that are challenging to solve, revealing the feature patterns responsible for this difficulty.

\begin{figure}[tbp]
    \centering
    \begin{subfigure}[b]{0.4\textwidth}
       \includegraphics[width=\linewidth]{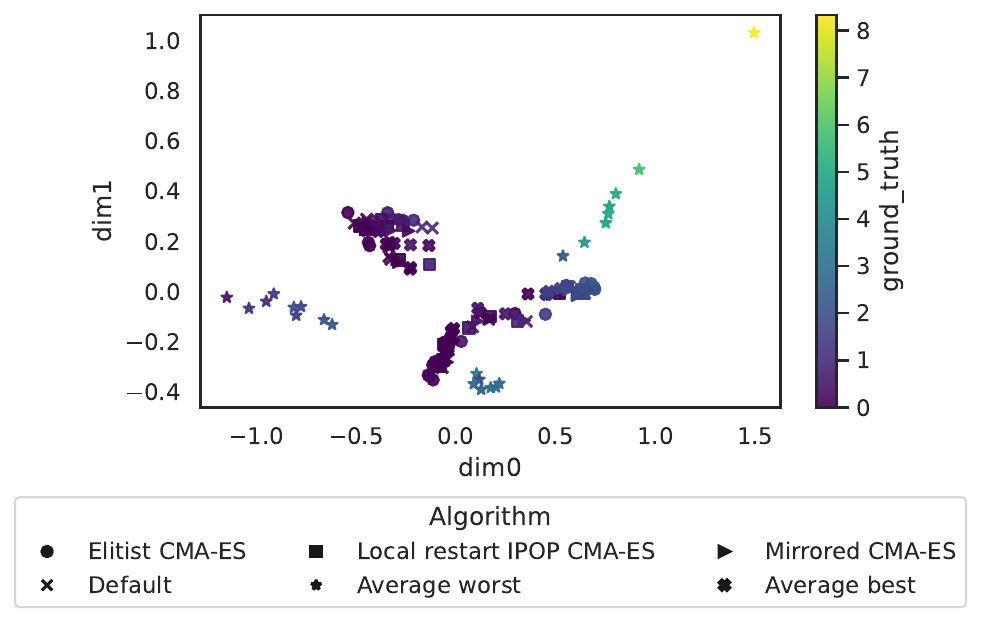}
       \caption{}
        \label{fig:ground_truth_2D_30d_modcma}
    \end{subfigure}
    \begin{subfigure}[b]{0.35\textwidth}
        \includegraphics[width=\linewidth]{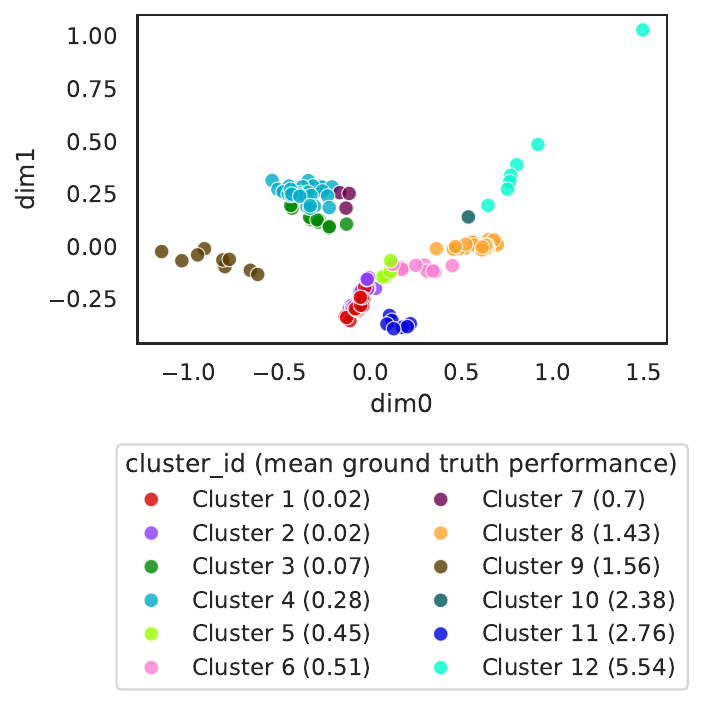}
        \caption{}
        \label{fig:clusters_2D_30d_modcma}
    \end{subfigure}
    \caption{The problem instances represented by their SHAP meta-features are projected into a two-dimensional vector space, with colors denoting: (a) the true performance, and (b) the cluster assignments of the SHAP meta-features for the thirty-dimensional (30$d$) test instances.}
    \label{fig:results_modcma_30d}
\end{figure}

\begin{figure*}[!htbp]
    \centering
    \includegraphics[trim=0.0cm 0.0cm 0.0cm 0.0cm, clip=true, scale=0.15]{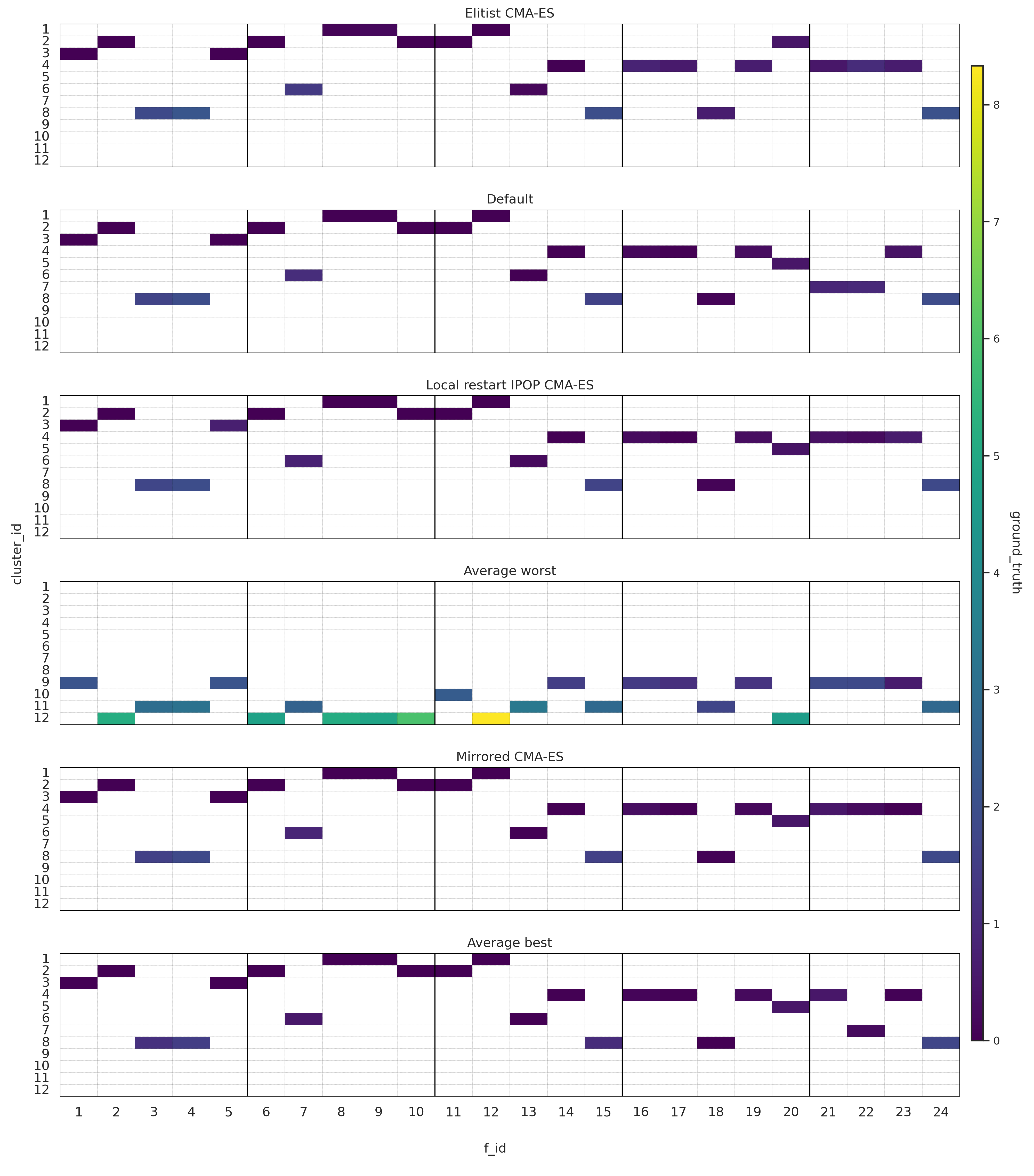}
    \label{fig:contagency_matrix_modcma_30d}
    \caption{Displayed are the footprints of the 30$d$ instances for each modCMA configuration, with footprint regions on the vertical axis and problem classes on the horizontal axis.}
    \label{fig:results_modcma_30d_footprint}
\end{figure*}

Figures \ref{fig:results_modcma_30d} and \ref{fig:results_modcma_30d_footprint} illustrate the performance of the modCMA configurations in 30-dimensional space. In particular, Figure \ref{fig:results_modcma_30d} is divided into two panels: (a) the ground-truth performance (Figure \ref{fig:ground_truth_2D_30d_modcma}), and (b) the clustering results of the problem instance SHAP meta-features (Figure \ref{fig:clusters_2D_30d_modcma}). The clustering identifies 12 distinct performance regions, shaped by diverse interactions among landscape features (Figure~\ref{fig:clusters_2D_30d_modcma}).

As in the 5$d$ case, most problem instances are effectively solved by all modCMA configurations (see the footprint in Figure~\ref{fig:results_modcma_30d_footprint}), except for those that pose challenges to the weakest performer (as indicated by the color map, where lower values represent better performance). Analysis on selected problems have been omitted due to the page limit.



Although the 5$d$ and 30$d$ footprints exhibit commonalities, these patterns are driven by each algorithm’s actual performance, good in lower dimensions does not necessarily carry over to higher ones.

\section{Discussion}
\label{sec:discussion}


By projecting various modular configurations into a joint space that captures the influence of landscape features on the performance of modCMA configurations, we can automatically identify distinct performance regions driven by different feature interactions. This analysis offers the advantage of avoiding subjective a priori thresholds to distinguish between good and poor performance. Instead, it uncovers all potential performance regions based on feature interactions, with performance gradually transitioning from best to worst.

By examining how the same problem instances are distributed across clusters under different configurations, we can determine which module combinations produce similar results (instances grouped in the same cluster) or different results (instances assigned to different clusters). Next, we can link these results to previous studies focusing solely on module contributions in the performance space, such as~\cite{nikolikj2024quantifying,van2024explainable}, establishing a connection between the landscape and the module parameter space. Additionally, the landscape features relevant to each set of modules can be identified and analyzed for each problem individually.

Footprints can be computed at any stage of the optimization run, making it possible to compare different evaluation budgets and identify which landscape features are most influential at each point. They also extend naturally to fixed-target settings, where performance is defined by the number of evaluations needed to reach a specified goal instead of by proximity to the optimum.

That said, although multi-target regression is in theory capable of modeling many targets, its predictive power diminishes when the number of targets (i.e., algorithms) greatly exceeds the number of available features. This challenge remains open in the ML field, but with more configurations and larger problem sets (e.g., the MA-BBOB benchmark suite~\cite{vermetten2025ma}), deep learning techniques could improve the MTR model to address a larger number of configurations.

Another limitation is the lack of diversity in the portfolio of six modCMA configurations studied, as the results indicate that five exhibit similar performance, with only one showing distinct behavior. In future work, we aim to perform a more in-depth analysis by employing automated methods~\cite{kostovska2023ps} to select a complementary portfolio of diverse algorithm configurations to be analyzed.

Finally, we did not consider alternative clustering methods or clustering quality measures. Future work could explore different clustering strategies to enable a sensitivity analysis of the clustering step.

\section{Conclusion}
\label{sec:conclusion}
This work explores the utility of algorithm footprints in understanding the dynamic interplay between algorithm configurations and problem characteristics. Through the analysis of six modular CMA-ES (modCMA) variants on 24 BBOB benchmark problems, evaluated in both 5-dimensional and 30-dimensional settings, we uncovered key insights into performance variability and the problem-specific features driving it. The study revealed both shared behavioral patterns across configurations, stemming from common interactions with problem properties, and distinct behaviors on the same problem, influenced by differing features. As part of our future work, we aim to perform an in-depth analysis of each problem on an individual basis, using various modular frameworks such as modCMA, modDE, and PSO-X.

\section*{Acknowledgment}
We gratefully acknowledge the Slovenian Research and Innovation Agency for funding through program grant P2-0098, project grants J2-4460 and GC-0001, as well as the young researcher grant PR-12897 awarded to AN. This research also received support from the European Union under Grant Agreement No. 101187010 (HE ERA Chair AutoLearn-SI). M.A. Muñoz’s contribution was funded by the Australian Research Council under grant IC200100009.


\end{document}